%% file: iclr2025_conference.tex
\documentclass{article} 
\usepackage{iclr2025_conference,times}

\input{math_commands.tex}

\usepackage{hyperref}
\usepackage{xurl}
\usepackage{xspace}

\usepackage{booktabs}       
\usepackage{amsfonts}       
\usepackage{nicefrac}       
\usepackage{microtype}      
\usepackage{xcolor}         
\usepackage{xspace}
\usepackage{natbib}

\usepackage{graphicx}
\usepackage{pifont}
\usepackage{tcolorbox}
\usepackage{tabularx}
\usepackage{makecell}
\usepackage{multirow}
\usepackage{multicol}
\usepackage{amssymb}
\usepackage{bbm}
\usepackage{amsmath}
\usepackage{wrapfig}
\usepackage{cleveref}

\newcommand\ours{\textsc{AutoTriton}\xspace}

\title{\ours: Automatic Triton Programming with Reinforcement Learning in LLMs}

\iclrfinalcopy

\author{%
Shangzhan Li$^\text{1,2}$,
\enspace Zefan Wang$^\text{1}$,
\enspace Ye He$^\text{1,2}$,
\enspace Yuxuan Li$^\text{1,}$\footnotemark[2],
\enspace Qi Shi$^\text{1,}$\footnotemark[2],
\enspace Jianling Li$^\text{3}$,
\\
\textbf{Yonggang Hu$^\text{4}$, 
\enspace Wanxiang Che$^\text{2}$,
\enspace Xu Han$^\text{1}$,
\enspace Zhiyuan Liu$^\text{1}$,
\enspace Maosong Sun$^\text{1}$
}\\[2pt]
\textsuperscript{1}Tsinghua University
\textsuperscript{2}Harbin Institute of Technology
\textsuperscript{3}Tianjin University \textsuperscript{4}OpenBMB\\[2pt]
\texttt{szli@ir.hit.edu.cn,  yxuanl1995@gmail.com, qshi9510@gmail.com} \\
\vspace{1mm}
}

\begin{document}

\maketitle

\begin{abstract}
  Kernel development in deep learning requires optimizing computational units across hardware while balancing memory management, parallelism, and hardware-specific optimizations through extensive empirical tuning. 
  Although domain-specific languages like Triton simplify GPU programming by abstracting low-level details, developers must still manually tune critical parameters such as tile sizes and memory access patterns through iterative experimentation, creating substantial barriers to optimal performance and wider adoption.  
  In this work, we introduce \ours, the first model dedicated to Triton programming powered by reinforcement learning (RL). 
  \ours performs supervised fine-tuning (SFT) to be equipped with essential Triton programming expertise using a high-quality data gathering pipeline, and conducts RL with Group Relative Policy Optimization (GRPO) algorithm, combining a rule-based reward and an execution-based reward to further improve Triton programming ability, sequentially. 
  Experiments across five evaluation channels of \textsc{TritonBench} and \textsc{KernelBench} illustrate that our $8$B model \ours achieves performance comparable to mainstream large models, including Claude-4-Sonnet and DeepSeek-R1-0528.
  Further experimental analysis demonstrates the crucial role of each module within \ours, including the SFT stage, the RL stage, and the reward design strategy. 
  These findings underscore the promise of RL for automatically generating high-performance kernels, and since high-performance kernels are core components of AI systems, this breakthrough establishes an important foundation for building more efficient AI systems. 
  The model and code will be available at \url{https://github.com/AI9Stars/AutoTriton}.
\end{abstract}

\renewcommand{\thefootnote}{\fnsymbol{footnote}}
\footnotetext[2]{Corresponding authors.}
\renewcommand{\thefootnote}{\arabic{footnote}}

\section{Introduction}
Efficient kernel engineering serves as the bedrock of high-performance deep learning systems, enabling models to execute optimally across an increasingly heterogeneous hardware landscape \citep{abadi2016tensorflow,paszke2019pytorch}. Historically, crafting such kernels in low-level languages like CUDA has been the exclusive domain of performance engineers, demanding intimate knowledge of hardware architecture and complex parallel programming patterns \citep{tillet2019triton}. The advent of Pythonic GPU programming frameworks, most notably Triton \citep{tillet2019triton}, has marked a significant leap in programmability. Notwithstanding these advances, such high-level abstractions have not fully eliminated the complexities of performance tuning. Developers are still burdened with the manual configuration of crucial parameters like tiling configurations and data layouts, a process of empirical trial-and-error that represents a primary bottleneck to realizing performance portability and widespread adoption.

\begin{figure}
  \centering
  \includegraphics[width=0.95\linewidth]{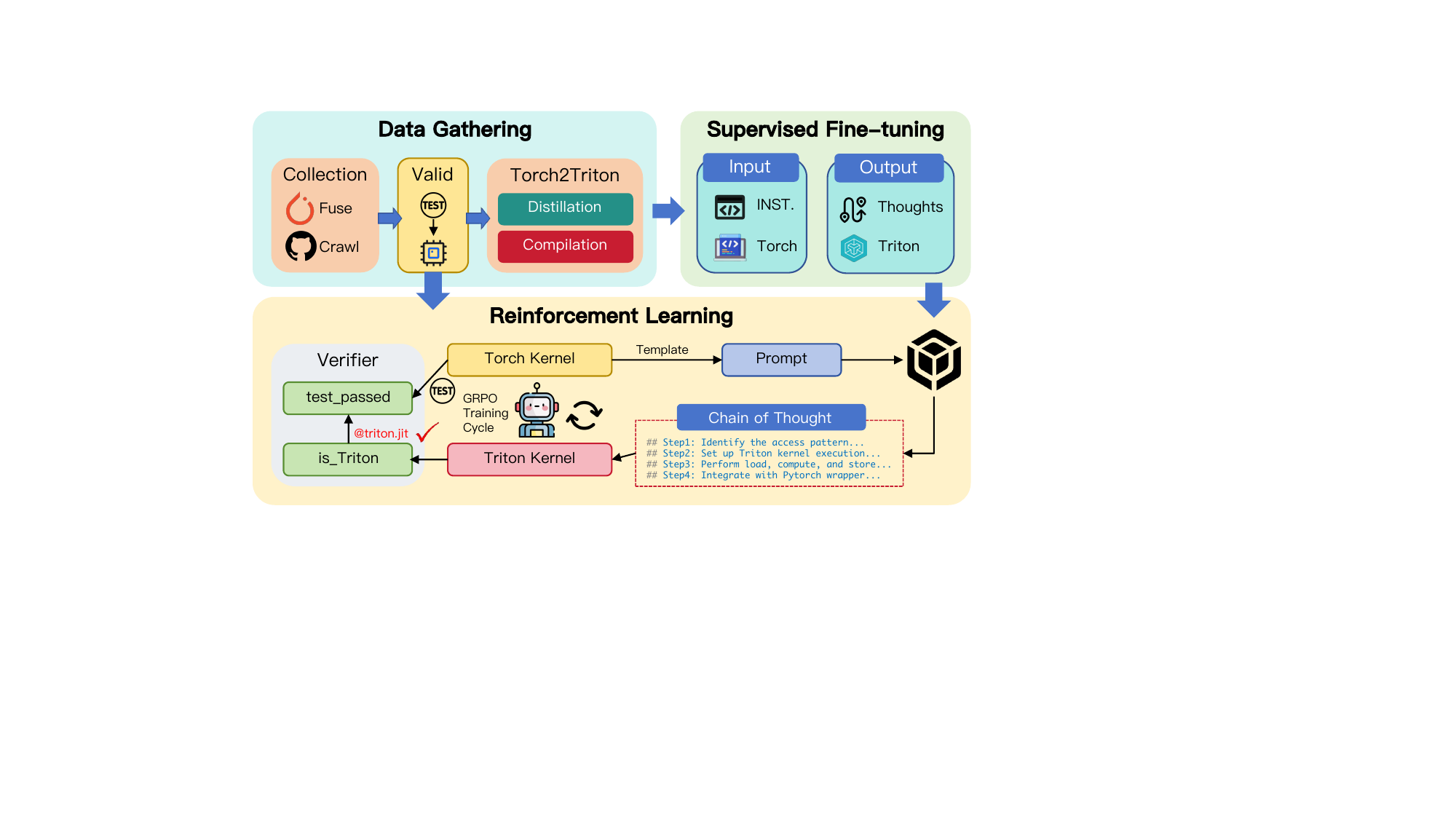}
  \caption{Overview of \ours pipeline. The entire pipeline consists of three components: data collection, SFT stage, and RL stage.}
  \vspace{-15pt}
  \label{fig:autotriton_pipeline}
\end{figure}

Current research in AI-assisted kernel generation has attracted increasing attention. Several benchmarks, such as \textsc{TritonBench} \citep{li2025tritonbench} and \textsc{KernelBench} \citep{ouyang2025kernelbench}, have been introduced to systematically evaluate the capabilities of LLMs in generating high-performance kernels. 
In addition to benchmarks, recent work such as AI CUDA Engineer \citep{lange2025ai} has gained widespread interest. This framework leverages general-purpose LLMs as foundation components to construct an automated workflow. However, its adaptability and flexibility remain limited due to the inherent capability boundaries of the underlying models.

In this work, we introduce \ours, the first model dedicated to Triton programming powered by reinforcement learning (RL).
\ours is built upon Seed-Coder-8B-Reasoning \citep{zhang2025seed}, which is a reasoning model dedicated to programming, further enhanced through a synergistic combination of supervised fine-tuning (SFT) and RL, which is shown in Figure~\ref{fig:autotriton_pipeline}.
In the SFT phase, we first design and implement a dedicated data construction pipeline. This pipeline is instrumental in assembling a high-quality Triton dataset that explicitly elucidates key programming concepts and reasoning steps inherent to Triton, thereby equipping \ours with foundational programming capabilities.
Subsequently, we leverage the data generated from the pipeline again and conduct RL with a combined rule-based and execution-based reward. This phase encourages the model to explore and internalize effective Triton programming strategies, allowing it to capture practical nuances and efficiencies that are challenging to instill through supervised fine-tuning alone.

Experimental results on two typical benchmarks \textsc{TritonBench} and \textsc{KernelBench} show that \ours achieves performance comparable to mainstream large models, including GPT-4o, Claude-4-Sonnet, Qwen3-32B, DeepSeek-R1-0528, on all five benchmark channels with only $8$B parameters, which indicates the effectiveness of \ours on the Triton programming task and highlights the crucial impact of our proposed data gathering pipeline and RL training strategy.
Further analysis underscores the pivotal roles of the SFT, RL, and reward design components in \ours. These findings offer what we consider to be crucial guidance for future research in this direction.

\section{Related Work}

\subsection{LLM for Kernel Generation}
Computation kernel generation is crucial for optimizing AI workloads on diverse hardware. Typical approaches, including MLIR \citep{mlir_llvm_project}, TVM \citep{apache_tvm}, collectively enhance AI model performance and portability, addressing the complexities of modern heterogeneous computing environments \citep{aldujaili2024looper}.
Recently, the automation of GPU kernel generation, critical for optimizing machine learning performance, has attracted significant research attention. Systematic evaluation of LLMs in this domain is facilitated by benchmarks such as \textsc{KernelBench} \citep{ouyang2025kernelbench}, which assesses the generation of fast and correct kernels across diverse workloads using metrics like ``fast\_p''.
While frontier models excel at general programming tasks, they often fall short on kernel generation tasks, underscoring a gap between general coding capabilities and specialized kernel optimization demands.
Similarly, \textsc{TritonBench} \citep{li2025tritonbench} highlights the challenges LLMs face with domain-specific languages like Triton, revealing difficulties in generating efficient kernels due to unfamiliarity with Triton's specifications and GPU programming intricacies.

Beyond benchmarks, frameworks like AI CUDA Engineer \citep{lange2025ai} utilize agentic approaches, leveraging LLMs for PyTorch-to-CUDA translation and iterative optimization. Despite achieving notable speedups, these training-free approaches are fundamentally constrained by the inherent limitations of the foundation LLMs. To directly enhance model capabilities, Kevin-32B \citep{baronio2025multi} employs multi-turn reinforcement learning, enabling the model to learn from environmental feedback and significantly improve kernel correctness and performance through self-refinement, particularly on complex tasks. Furthermore, the DeepSeek-R1 model, augmented with test-time scaling, demonstrates the efficacy of allocating increased inference compute for iterative refinement and verification, achieving high correctness on \textsc{KernelBench} tasks \citep{nvidia2025deepseek_kernelgen}. These advancements collectively indicate a trend towards iterative, feedback-driven methodologies to enhance LLM proficiency in specialized high-performance code generation.
Additionally, \textsc{KernelLLM} \citep{kernelllm2025} generates Triton kernels via supervised fine-tuning. Despite achieving reasonable performance, it is fundamentally constrained by the ceiling of imitation learning. It fails to leverage exploration, thereby limiting its ability to produce higher-quality Triton kernels.
Different from the above works, in this work, we propose \ours, the first model specifically designed for Triton programming with reinforcement learning, achieving remarkable improvements across five typical benchmark channels.

\subsection{RL for Code}
RL provides a powerful paradigm for agents to learn optimal policies through interaction with dynamic environments, maximizing cumulative rewards.
Early applications in code generation formulate the problem within a Markov Decision Process (MDP), where partial programs constitute states and grammar productions serve as actions \citep{Concord2020}. This formulation highlights RL's flexibility in adapting to different levels of abstraction. Modern advancements leverage LLMs, treating the code-generating model as an actor and code generation as actions, with functional correctness derived from unit test results providing the reward signal \citep{CodeRL2022}. This approach has enabled systems like AlphaCode to achieve competitive performance in complex coding tasks \citep{AlphaCode2022}. Beyond generation, RL is extensively applied in code optimization, notably for learning optimal sequences of compiler passes \citep{SurveyRLCodeLLMs, Shahzad2022}. Here, states are often represented by statistical analyses of Intermediate Representation (IR) or graph-based models, and rewards are tied to performance metrics such as cycle count, area, or resource utilization \citep{Shahzad2022}. The success of frameworks like CYCLE further demonstrates RL's capacity for iterative self-refinement of faulty code generations, learning from execution feedback, and significantly improving refinement capabilities \citep{CYCLE2024}. 

Despite these advancements, RL for code faces substantial challenges. 
Designing robust reward functions remains a primary concern. Poorly engineered rewards can lead to unintended behaviors or "reward hacking", where the agent exploits the reward structure rather than achieving the intended objective \citep{MilvusRLLimitations}. 
Training instability, especially when fine-tuning large language models, presents another hurdle, with algorithms like REINFORCE++ \cite{hu2025reinforce++} often suffering from volatile policy updates. To address these instabilities and enhance training convergence, improved algorithms such as Group-Relative Policy Optimization (GRPO) have been developed, which also help in eliminating reward hacking within RL frameworks for LLMs \citep{shao2024deepseekmath}. 
Our proposed \ours aligns with this perspective. In the field of kernel generation, we employ the GRPO algorithm and design reasonable rewards to build the LLM’s understanding of kernel queries and conduct RL-powered Triton programming accordingly.

\section{\ours}
In this section, we introduce \ours, a specialized model tailored for the Triton programming task.
\ours is characterized by a sequential two-stage process.
Initially, the model undergoes SFT to establish a strong foundation in Triton programming principles. Following this, an RL framework is applied, which allows execution-based feedback of GPU code, guiding the model to further optimize the generated kernels' correctness. 
To elucidate \ours, we will first formally perform the task formulation, then detail the supervised fine-tuning procedure, and subsequently present the design of the reinforcement learning framework.

\subsection{Problem Formulation}
Developing custom kernels traditionally demands substantial domain expertise and involves a significant amount of empirical trial-and-error. 
To accelerate this development lifecycle, we define the task of Triton programming. 
This task aims to learn a mapping from a comprehensive kernel specification to its corresponding executable Triton implementation.
A kernel specification $\mathcal{D}$, generally comprises two primary components: a concrete PyTorch implementation or a formal interface definition that details its functional description, the signatures of input and output parameters (including data types), and the dimensionality (shapes) of the tensors involved. The core challenge is to develop a model $\mathcal{M}$ that, given $\mathcal{D}$, synthesizes a Triton kernel $\mathcal{T}$ that is not only syntactically correct and executable but also semantically faithful to all requirements outlined in $\mathcal{D}$.

\subsection{Supervised Fine-tuning}
\label{sec:sft}
\begin{figure}
  \centering
  \includegraphics[width=0.97\linewidth]{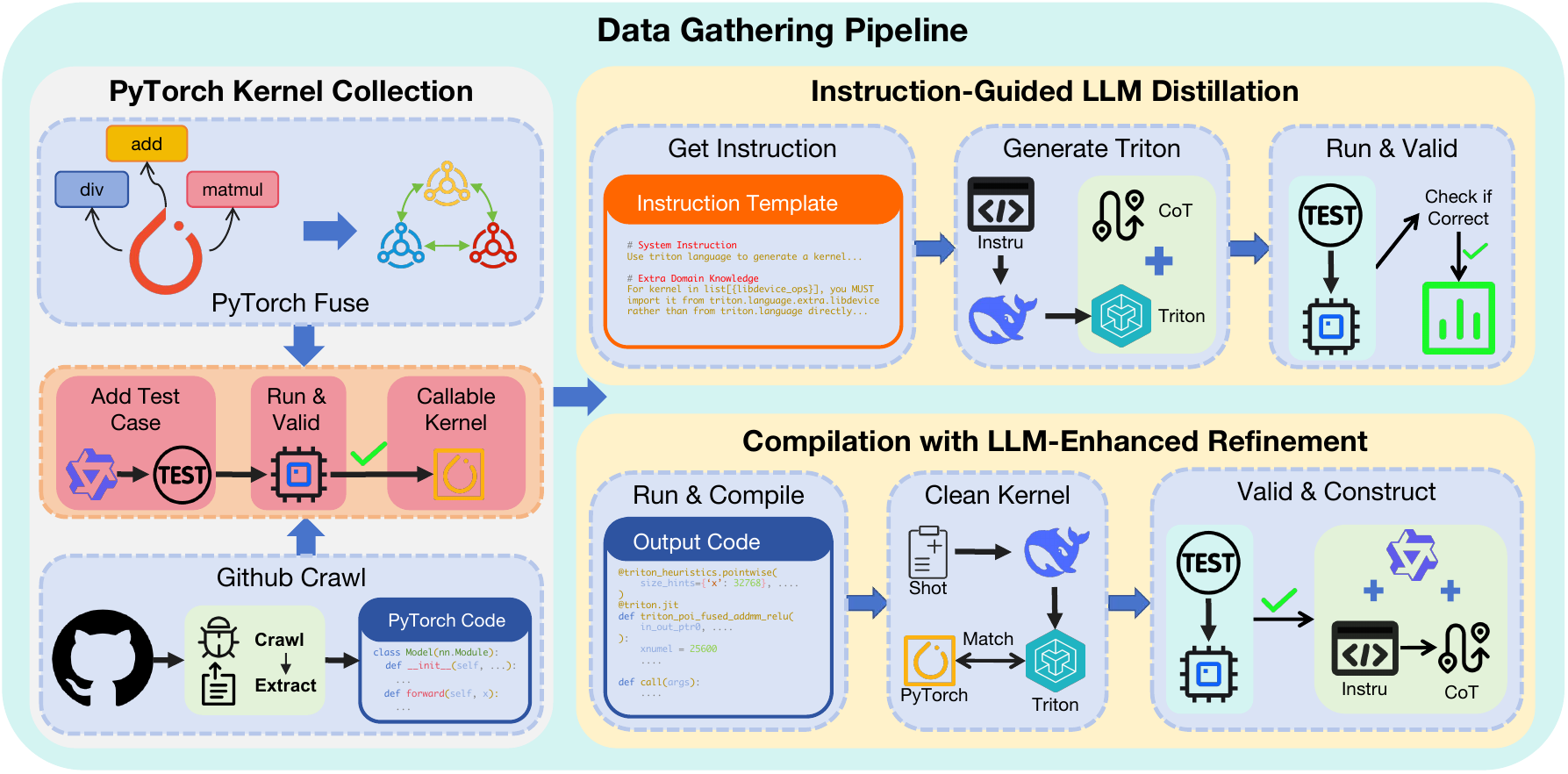}
  \caption{Data gathering pipeline of \ours. Our pipeline begins with the \textbf{systematic collection of PyTorch kernels}, then generates corresponding Triton kernels by \textbf{instruction-guided LLM distillation} and \textbf{compilation with LLM enhanced refinement} simultaneously.}
  \label{fig:data_pipeline}
\end{figure}

Recent studies \citep{li2025tritonbench} have highlighted that even models proficient in general-purpose programming exhibit limited capabilities in generating specialized Triton kernels.
To bridge this capability gap and equip our model with essential Triton programming expertise, we develop a meticulous data gathering pipeline to produce high-quality data for supervised fine-tuning.
This pipeline automates the crucial steps of data collection, synthesis, and validation, with the explicit goal of retaining only high-fidelity, syntactically sound, and demonstrably correctly executable data for training.
The architecture of the pipeline is illustrated in Figure \ref{fig:data_pipeline}. 

Our proposed pipeline for data acquisition begins with the \textbf{systematic collection of PyTorch kernels}. This entails harvesting kernels from established open-source platforms like GitHub and HuggingFace, supplemented by the algorithmic composition of basic kernels through the PyTorch interface. We then leverage an open-source LLM proficient in programming, such as Qwen 2.5 Coder \citep{hui2024qwen2}, for the automated generation of test cases, which are subsequently used to validate and retain executable PyTorch kernels.

Following the collection of PyTorch kernels, we employ two distinct strategies for generating corresponding Triton kernels: \textbf{instruction-guided LLM distillation} and \textbf{compilation with LLM-enhanced refinement}.

The distillation-based approach involves creating targeted instructions that encapsulate both the PyTorch kernel's functionality and relevant Triton-specific knowledge. A capable deep-reasoning LLM, such as DeepSeek R1 \citep{guo2025deepseek}, is prompted with these instructions to produce Triton code, accompanied by a step-by-step Chain-of-Thought (CoT) explanation. The generated Triton snippets are then cross-validated against the original PyTorch kernels using the previously generated test cases, and only functionally equivalent pairs are selected for supervised fine-tuning.

Recognizing the inherent limitations of general-purpose LLMs in proficiently generating Triton code \citep{li2025tritonbench}, we also leverage a compilation-based approach for enhanced data acquisition efficiency. 
Specifically, PyTorch code snippets are processed using \texttt{torch.compile}. The resultant compiled artifacts are then refined by an LLM to improve human readability; this involves tasks such as inserting explanatory comments, removing extraneous decorators, and renaming variables to be more semantically meaningful. 
After verifying functional equivalence with PyTorch using test cases, we leverage an LLM to craft instructions. These, along with the verified Triton code, are used to prompt an LLM to generate detailed CoT narratives, aiming to instill Triton programming paradigms during supervised fine-tuning.

Finally, the culminating dataset, comprising \texttt{<instruction, Triton code with CoT>} pairs, is leveraged for supervised fine-tuning. 
The model learns to predict the Triton code and its CoT justification conditioned on the instruction prompt. 
thereby developing foundational capabilities in Triton programming.

\subsection{Reinforcement Learning}
\label{sec:rl}
To further push the border of the coding ability of \ours, we adopt a common Reinforcement Learning with Verifiable Reward~(RLVR) pipeline. 
Our training process is based on the GRPO algorithm~\citep{shao2024deepseekmath}, updating the policy with a group normalized objective: 
\begin{equation}
\scalebox{0.76}{ 
  $\displaystyle 
  \begin{aligned}
  \mathcal{J}_{GRPO}(\theta) ={}& \mathbb{E}[q \sim P(Q), \{o_i\}_{i=1}^G \sim \pi_{\theta_{old}}(O|q)] \\
  & \frac{1}{G} \sum_{i=1}^G \frac{1}{|o_i|} \sum_{t=1}^{|o_i|} \left\{ \min \left[ \frac{\pi_\theta(o_{i,t}|q, o_{i,<t})}{\pi_{\theta_{old}}(o_{i,t}|q, o_{i,<t})} \hat{A}_{i,t}, \text{clip} \left( \frac{\pi_\theta(o_{i,t}|q, o_{i,<t})}{\pi_{\theta_{old}}(o_{i,t}|q, o_{i,<t})}, 1-\epsilon, 1+\epsilon \right) \hat{A}_{i,t} \right] - \beta D_{KL}(\pi_\theta || \pi_{ref}) \right\}
  \end{aligned}
  $
}
\end{equation}

where $\pi_\theta$ and $\pi_{\theta_{old}}$ are the policy model and reference model, and $\hat{A}_{i,t}$ is the group-wise advantage:
\begin{equation}
\scalebox{1.0}{
$\displaystyle
\begin{aligned}
\hat A_{i,t} = \frac{r_i-\text{mean}(\{r_j\}_{j=1}^N)}{\text{std}(\{r_i\}_{i=1}^N)}. 
\end{aligned}
$
}
\end{equation}

The reward function is defined by the following equation, which combines an execution-based component with a rule-based one:
\begin{equation}
R(\hat{a}) = \begin{cases} 
1, & \texttt{is\_Triton}(\hat{a}) \ \& \   \texttt{test\_passed}(\hat{a}) \\
0, & \text{otherwise} 
\end{cases}
\end{equation}
where the \texttt{test\_passed} component confirms functional correctness. It verifies that the generated code passes all test cases and is semantically equivalent to a reference PyTorch implementation, within a tolerance of $\epsilon$.
While the \texttt{is\_Triton} component ensures syntactic validity by using a rule-based linter to check if the code conforms to the Triton language specification. This component regularizes the policy to discourage "reward hacking," a scenario where the model might generate non-Triton code (such as simpler PyTorch code) that would trivially pass the functional tests.

The data for our RL stage is also generated using the pipeline from \S~\ref{sec:sft}. In this stage, we only retain \texttt{<instruction, PyTorch code>} pairs, as the reference PyTorch code is sufficient for deriving a reward signal through test-case execution, eliminating the need for labeled Triton code. This enables the inclusion of more difficult, out-of-distribution (OOD) data well-suited for RL exploration. The final training data is a strategic mix of these novel instances and a small portion of in-distribution data from the SFT phase to ensure a smooth policy transition.

\section{Experiments}
In this section, we evaluate the performance of \ours and conduct a comprehensive analysis from multiple aspects.

\subsection{Evaluation Setup}

\paragraph{Evaluation Benchmarks}

We evaluate \ours using two established benchmarks: \textsc{TritonBench} \citep{li2025tritonbench} \footnote{We use \textsc{TritonBench-T} version in \url{https://github.com/thunlp/TritonBench/pull/6}.} and \textsc{KernelBench} \citep{ouyang2025kernelbench} \footnote{We use the Triton backend version of \textsc{KernelBench} in \url{https://github.com/ScalingIntelligence/KernelBench/pull/35}.}. \textsc{TritonBench} assesses LLM capabilities in generating Triton kernels, which is divided into two evaluation channels: \textsc{TritonBench-G} consists of $184$ real-world kernels from GitHub and \textsc{TritonBench-T} consists of $166$ kernels aligned with PyTorch interfaces.
While \textsc{KernelBench} evaluates LLM proficiency in generating efficient GPU kernels for neural network optimization across $250$ tasks, categorized into Level $1$ ($100$ single-kernel tasks, e.g., convolution, for CUDA replacement), Level $2$ (100 simple fusion tasks, e.g., conv+bias+ReLU, for fused CUDA kernels), and Level $3$ ($50$ full architecture tasks, e.g., MobileNet, for end-to-end CUDA optimization). The prompts used during inference are detailed in figure~\ref{fig:infer_prompts}.

\paragraph{Evaluation Metrics}
Regarding the evaluation metrics, we synthesize those from the above two benchmarks and categorize them into two aspects: (1) \texttt{Compilation Accuracy} (error-free compilations);
(2) \texttt{Call Accuracy} (error-free invocation); (3) \texttt{Execution Accuracy} (correct input-output behavior); (4) \texttt{Speed Up} (relative execution time improvement) on both benchmarks.
Following the original settings of both benchmarks, we evaluate \texttt{Compilation Accuracy}, \texttt{Execution Accuracy}, \texttt{Speed Up} on \textsc{KernelBench}, and evaluate \texttt{Call Accuracy}, \texttt{Execution Accuracy}, \texttt{Speed Up} on \textsc{TritonBench} respectively.
For evaluations on \textsc{TritonBench-G}, the \texttt{Speed Up} value is derived by comparing against their supplied reference Triton code, while for all other evaluations, \texttt{Speed Up} is calculated against PyTorch implementations.
Following \textsc{KernelBench}, we report $\text{fast}_p$ to measure the absolute speedup of Triton codes across the entire benchmark, which is calculated as follows:
\begin{equation}
    \text{fast}_p = \frac{1}{N} \sum_{i=1}^{N} \mathbbm{1}(\text{correct}_i \land \{\text{SpeedUp}_i > p\}),
\end{equation}

\paragraph{Training Details}
In the SFT stage, we utilize the LLaMA-Factory framework \citep{zheng2024llamafactory} with a dataset of $14,102$ samples. We set the maximum sequence length to $16,384$ and use a training batch size of $1$ per device. The model is fine-tuned with a learning rate of $1 \times 10^{-5}$ for 3 epochs. This stage is completed in approximately $16$ hours on a single node with $8$ A800 GPUs.
For the subsequent RL stage, we adopt the VeRL framework  \citep{sheng2025hybridflow}, using the dataset containing $6,302$ samples. In this phase, the training batch size is set to $64$. The maximum prompt length is capped at $4,096$ tokens, while the maximum response length is set to $16,384$ tokens. The learning rate for the actor's optimizer is configured to $1 \times 10^{-6}$. The model is trained for $1$ epoch. This phase requires approximately $32$ hours of training time on two nodes, utilizing a total of $16$ A800 GPUs.

\subsection{Main Results}

Table~\ref{tab:tritonbench_result} and Table~\ref{tab:kernelbench_result} show the experimental results of \ours on \textsc{TritonBench} and \textsc{KernelBench}, respectively.
Experiments are systematically conducted across five evaluation channels of the \textsc{TritonBench} and \textsc{KernelBench} benchmarks.
In terms of correctness (\texttt{Exec} in Table), \ours decisively surpasses powerful models including DeepSeek-R1-0528, GPT-4o, Claude-4-Sonnet, DeepSeek-R1-0120, and Qwen3-32B, 
which proves the effectiveness of \ours across both its SFT and RL phases, underscoring the significant contributions of our proposed data gathering pipeline and training framework.
The superiority of \ours is also evident in the runtime performance evaluation. It achieves performance comparable to mainstream large models, including claude-4-sonnet and DeepSeek-R1-0528, on most evaluation channels, underscoring the effectiveness of our proposed data gathering pipeline in yielding high-fidelity training instances.

It is further observed that on the \textsc{TritonBench-G} channel, all evaluated models struggle significantly on both correctness and performance metrics.
The inherent difficulty of this channel, which evaluates models on real-world requirements from GitHub against reference Triton implementations, highlights the substantial challenges that persist in automated Triton programming. This suggests the task is far from solved and warrants deeper investigation.

\input{tables/main_result_tritonbench}

\input{tables/main_result_kernelbench}

\subsection{Analysis}
\paragraph{Cross Comparisons for Triton and CUDA Models}
To further assess the performance of \ours, we conduct a comparative analysis against prominent kernel generation models not specifically focused on Triton programming, namely AI CUDA Engineer \cite{lange2025ai} and Kevin-32B \cite{baronio2025multi}.
We select the \textsc{KernelBench} benchmark for this evaluation due to its capability to assess both Triton and CUDA kernels, ensuring a fair comparison.
As illustrated in Table~\ref{tab:pass@10_result}, we report the P$75$ and P$50$ speedups over the PyTorch baseline in the pass@10 setting, which represent the speedup ratios at the $75$th and $50$th percentiles of the kernel performance distribution. 

A key finding from our analysis is the persistent, systematic gap in automated programming proficiency between Triton and CUDA, which highlights the formidable challenges associated with high-performance Triton code generation. Even against this backdrop, \ours establishes its superiority over recent specialized framework AI Cuda Engineer \citep{lange2025ai}, delivering quantitatively better results on metrics of both correctness and runtime efficiency. While these results validate the strength of \ours, it still lags behind the Kevin-32B \citep{baronio2025multi} model, which we attribute to three potential causes: the intrinsic programming model differences between CUDA and Triton, a parameter scale mismatch ($8$B vs. $32$B), and the use of $90$\% of the evaluation data in Kevin-32B's training, which likely results in higher evaluation scores.

\input{tables/pass_10_results}

\paragraph{Effects of Reinforcement Learning}

The final rows of Table~\ref{tab:tritonbench_result} and Table~\ref{tab:kernelbench_result} present the performance of \ours without the RL stage. A clear performance uplift is observed when comparing \ours to its SFT-only counterpart, demonstrating that RL is effective at raising the performance ceiling for the Triton programming task. This result suggests that RL enables the model to transcend the inherent limitations of imitation learning, which aligns with findings across numerous other domains.
Furthermore, the performance gains achieved during the RL stage also validate the efficacy of our proposed data gathering pipeline. This pipeline effectively generates a training dataset that is highly suitable for exploration during RL, which plays a crucial role in pushing the boundaries of the Triton programming task.

\paragraph{Effects of Reward Design}
\input{tables/reward_design_role}

\begin{figure}
  \centering
  \includegraphics[width=0.97\linewidth]{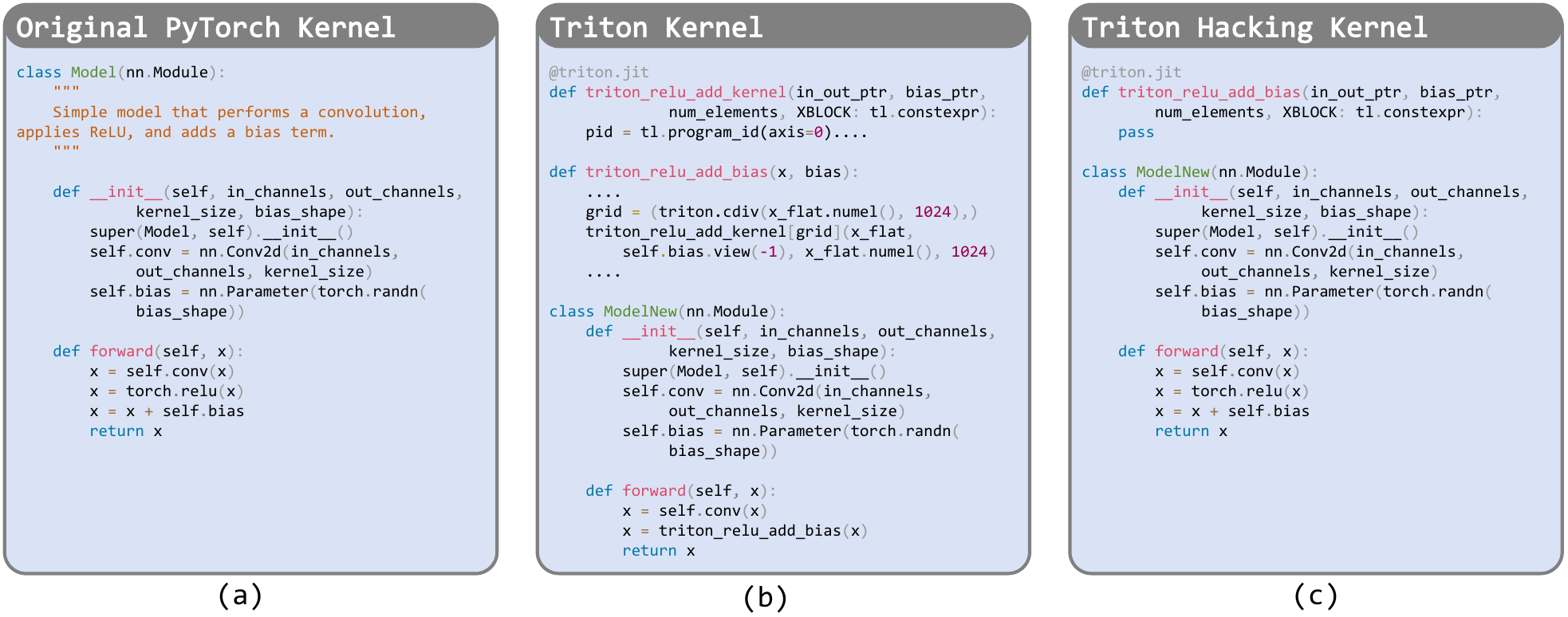}
  \caption{Example of the phenomenon of the low-quality implementation of Triton code.}
  \label{fig:case_studies}
\end{figure}

As mentioned in \S~\ref{sec:rl}, a primary challenge in the Triton programming task is reward hacking, where models learn to satisfy test cases without generating correct Triton code. 
To address this, we introduce auxiliary rule-based rewards alongside the primary execution-based reward to explicitly incentivize adherence to the Triton language specification. 
The impact of this strategy is quantified in Table~\ref{tab:reward_design_role}. By checking for the mandatory "\texttt{@triton.jit}" decorator, we find that rule-based rewards significantly decrease the count of invalid generations on \textsc{TritonBench-T} (from $18$ to $5$) and \textsc{KernelBench}-Level1 (from $25$ to $6$), confirming the importance of explicit syntactic guidance in the reward mechanism.
Despite these improvements, a rule-driven reward function can still be hacked. Models may learn to generate low-quality code that satisfies the explicit rules but fails to fulfill the complete semantic requirement of Triton.
For instance, as illustrated in Figure~\ref{fig:case_studies}, when tasked with implementing a kernel composed of a convolution and a ReLU (Figure~\ref{fig:case_studies}(a)), the model often generates a valid Triton kernel for the simpler ReLU part while leaving the more complex convolution as a fallback PyTorch implementation (Figure~\ref{fig:case_studies})(b). 
More critically, the model might circumvent the reward rules entirely by generating a fake Triton kernel that it never calls (Figure~\ref{fig:case_studies})(c).
This phenomenon of low-quality implementation is highly prevalent across all evaluation models. 
A potential countermeasure involves incorporating runtime-based performance rewards, which we reserve for future work.

\begin{figure}
  \centering
  \includegraphics[width=0.5\linewidth]{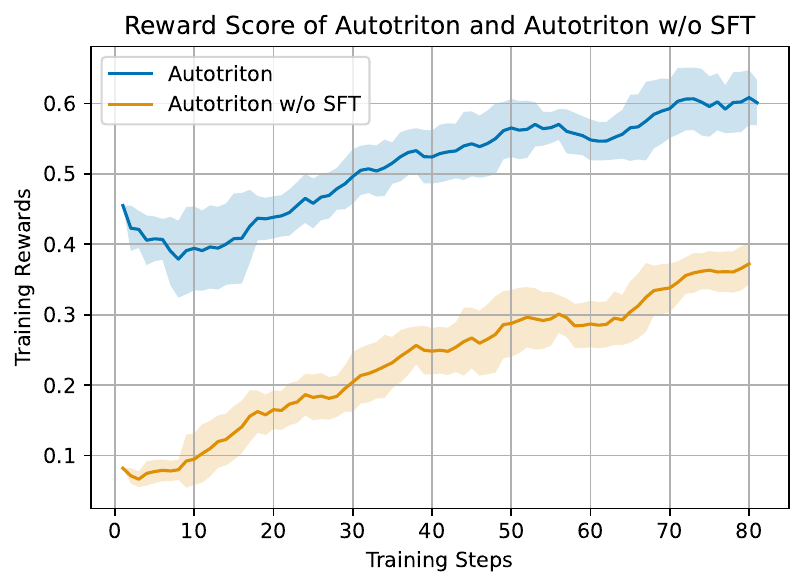}
  \caption{Reward scores of \ours and \ours w/o SFT stage.}
  \label{fig:reward_score}
\end{figure}

\paragraph{Effects of Supervised Fine-tuning}
As shown in Table~\ref{tab:tritonbench_result} and Table~\ref{tab:kernelbench_result}, after undergoing SFT, the model achieves superior performance compared to the original backbone model (Seed-Coder-8B-Reasoning). This initial result suggests that SFT is effective in familiarizing the model with the fundamental paradigms of Triton programming and further proves the effectiveness of our proposed data gathering pipeline in generating high-quality SFT data.
This conclusion is further substantiated by the training dynamics in Figure~\ref{fig:reward_score}. 
Although the model without SFT also exhibits a notable upward trend in its reward curve, it suffers from severe reward hacking, with the majority of instances displaying the behavior illustrated in Figure~\ref{fig:case_studies}(c).
Specifically, while its generated code can pass the test cases, they often fail to adhere to Triton's basic syntax, defaulting instead to simpler Torch implementations, which is consistent with the phenomenon observed in Table~\ref{tab:reward_design_role}.
This behavior indicates that SFT is crucial not only for learning the correct syntax but also for preventing reward hacking, where the model learns to exploit test cases with trivial Torch code rather than mastering genuine Triton programming.

\paragraph{Limitations}
Based on the multi-faceted evaluation of \ours, the primary limitation is that our current training framework \textbf{lacks performance-guided training}. 
Since the compiled or distilled kernels lack efficient runtime feedback,
\ours does not contain performance-guided training in the current stage.
Future work could focus on procuring or generating higher-quality data and integrating a performance-aware training framework simultaneously, rewarding the model for generating kernels that are not only functionally correct but also achieve high performance on target hardware, thereby guiding it toward more efficient solutions.

\section{Conclusion}
In this work, we study the task of automated Triton programming and propose \ours, the first model dedicated to Triton programming powered by RL. 
\ours involves a two-stage training process: an SFT stage where \ours learns essential Triton programming expertise from high-quality data generated by our novel data curation pipeline, followed by an RL stage where it further improves by exploring more challenging problem instances.
Evaluations across five channels of \textsc{TritonBench} and \textsc{KernelBench} show that \ours achieves performance comparable to mainstream large models, including claude-4-sonnet and DeepSeek-R1-0528. 
Our in-depth analysis of each component validates the significant potential of RL-based methods for automatic Triton programming. 
Ultimately, \ours demonstrates a promising pathway toward the automated generation of efficient kernels, offering a new paradigm for building high-performance AI systems.

\section*{Acknowledgements}
The work is initiated and supported by the AI9Stars Team. We are grateful for the support of the OpenBMB and InfiniteTensor teams.

\bibliography{iclr2025_conference}
\bibliographystyle{iclr2025_conference}

\appendix
\section{Inference Prompts}

\begin{figure}[h]
  \centering
  \includegraphics[width=0.9\linewidth]{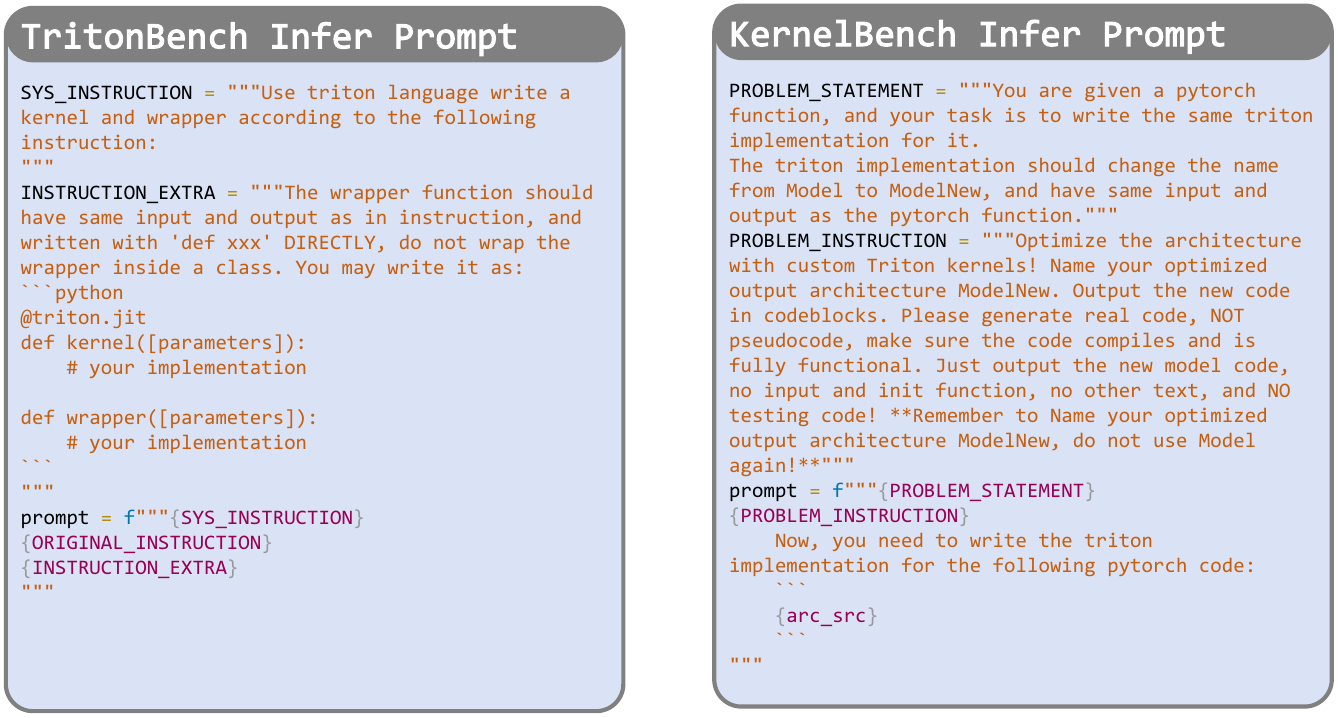}
  \vspace{-5pt}
  \caption{\ours prompts for experimental reasoning.}
  \vspace{-15pt}
  \label{fig:infer_prompts}
\end{figure}

\end{document}

%% file: math_commands.tex
\usepackage{amsmath,amsfonts,bm}

\def\eqref#1{equation~\ref{#1}}

\def\1{\bm{1}}

\DeclareMathAlphabet{\mathsfit}{\encodingdefault}{\sfdefault}{m}{sl}
\SetMathAlphabet{\mathsfit}{bold}{\encodingdefault}{\sfdefault}{bx}{n}



%% file: tables/main_result_tritonbench.tex
\begin{table*}[ht!] \small
    \centering
    \resizebox{0.75\textwidth}{!}{
        \begin{tabular}{lccccccc}
        \toprule

        \multirow{2}{*}{\textbf{Model}} 
        & \multirow{2}{*}{\textbf{\#Params}}
        & \multicolumn{2}{c}{\textbf{\textsc{TritonBench-G}}} & \multicolumn{2}{c}{\textbf{\textsc{TritonBench-T}}} \\
        \cmidrule(lr){3-4}
        \cmidrule(lr){5-6}

        & & \textbf{Call}~/~\textbf{Exec} & $\textbf{fast}_\textbf{1}$~/~$\textbf{fast}_\textbf{2}$ & \textbf{Call}~/~\textbf{Exec} & $\textbf{fast}_\textbf{1}$~/~$\textbf{fast}_\textbf{2}$   \\         
        \midrule
        Seed-Coder-Reasoning       &  $8$B    & $2.72$~/~$2.72$ & $0.54$~/~$0.00$                         &$3.61$~/~$3.61$ & $1.20$~/~$0.00$ \\
        Qwen3       &    $8$B  & $2.17$~/~$1.63$       &$0.54$~/~$0.00$ &$6.02$~/~$5.42$   & $2.41$~/~$0.00$   \\
        Qwen3      &     $32$B       & $10.33$~/~$9.24$      &$2.17$~/~$\underline{1.63}$      &$21.96$~/~$21.96$   &$10.84$~/~$3.01$      \\
        GPT-4o     & -	& $10.87$~/~$10.33$   & $\underline{4.89}$~/~$\underline{1.63}$  & $18.67$~/~$15.06$   &$7.84$~/~$1.20$  \\
        Claude-4-Sonnet    & -  & $9.24$~/~$9.24$                &$1.64$~/~$1.09$ & $10.84$~/~$10.84$  &$4.22$~/~$1.84$ \\
        DeepSeek-R1-0120     & $671$B  & $13.59$~/~$13.05$  &$\underline{4.89}$~/~$0.54$      & $28.92$~/~$28.37$     &$\mathbf{22.89}$~/~$3.01$    \\
        DeepSeek-R1-0528      &    $685$B        &$\mathbf{16.30}$~/~$\underline{15.22}$ &$3.80$~/~$0.54$             &$30.72$~/~$30.12$           & $11.45$~/~$4.82$       \\
        \ours  & $8$B & $\underline{15.76}$~/~$\mathbf{15.76}$  &$\mathbf{7.61}$~/~$\mathbf{2.17}$       &$\mathbf{40.36}$~/~$\mathbf{39.16}$   &$\underline{17.04}$~/~$\underline{6.02}$ \\
        \ \ \ w/o RL (SFT only)  & $8$B & $14.67$~/~$14.13$ &$\underline{4.89}$~/~$1.09$&$\underline{34.94}$~/~$\underline{34.94}$   &$15.06$~/~$\mathbf{7.83}$      \\
        \bottomrule
    \end{tabular}
    }
    \caption{Main results on \textsc{TritonBench}. 
    We present \texttt{Call Accuracy} (Call), \texttt{Execution Accuracy} (Exec), $\text{fast}_\text{1}$ and $\text{fast}_\text{2}$.
    The best-performing and second-best-performing methods are highlighted in \textbf{Bold} and \underline{Underline}, respectively.}
    \label{tab:tritonbench_result}
\end{table*}

%% file: tables/main_result_kernelbench.tex
\begin{table*}[ht!] \small
    \centering
    \resizebox{0.99\textwidth}{!}{
        \begin{tabular}{lcccccccccc}
        \toprule

        \multirow{2}{*}{\textbf{Model}} 
        & \multirow{2}{*}{\textbf{\#Params}} 
        & \multicolumn{2}{c}{\textbf{\textsc{Level1}}} & \multicolumn{2}{c}{\textbf{\textsc{Level2}}}  & \multicolumn{2}{c}{\textbf{\textsc{Level3}}} \\
        \cmidrule(lr){3-4}
        \cmidrule(lr){5-6}
        \cmidrule(lr){7-8}

        & &  \textbf{Comp}~/~\textbf{Exec} & $\textbf{fast}_\textbf{1}$~/~$\textbf{fast}_\textbf{2}$ & \textbf{Comp}~/~\textbf{Exec} & $\textbf{fast}_\textbf{1}$~/~$\textbf{fast}_\textbf{2}$ &\textbf{Comp}~/~\textbf{Exec} & $\textbf{fast}_\textbf{1}$~/~$\textbf{fast}_\textbf{2}$
        \\         
        \midrule
        Seed-Coder-Reasoning        &  $8$B   & $48.0$~/~$10.0$                & $4.0$~/~ $2.0$                      &$44.0$~/~$11.0$   & $5.0$~/~$\mathbf{4.0}$    &$52.0$~/~$10.0$  &$4.0$~/~ $\mathbf{4.0}$ \\
        Qwen3      &   $8$B    & $52.0$~/~$16.0$       &$9.0$~/~$\mathbf{9.0}$                        &$73.0$~/~$16.0$   & $8.0$~/~$1.0$     &$40.0$~/~$14.0$   &$4.0$~/~$2.0$   \\
        Qwen3       &  $32$B         & $84.0$~/~$23.0$       &$5.0$~/~$4.0$      &$\mathbf{98.0}$~/~$25.0$   &$15.0$~/ $2.0$    &$\mathbf{92.0}$~/~$16.0$  &$6.0$~/~$0.0$ \\
        GPT-4o  & - & $\underline{91.0}$~/~$15.0$  &$3.0$~/~$1.0$    &$83.0$~/~$5.0$   &$3.0$~/~$0.0$   &$74.0$~/~$8.0$ &$4.0$~/~$2.0$ \\
        Claude-4-Sonnet  & -  & $87.0$~/~$33.0$  &$\mathbf{11.0}$~/~$\underline{7.0}$ &$92.0$~/~$26.0$   &$10.0$~/~$1.0$   &$\underline{82.0}$~/~$18.0$ &$2.0$~/~$0.0$ \\
        KernelLLM & $8$B & $72.0$~/~$20.2$ & $-$~/~$-$ & $76.0$~/~$16.0$ & $-$~/~$-$ & $-$~/~$-$ & $-$~/~$-$ \\
        DeepSeek-R1-0120 & $671$B & $\mathbf{95.0}$~/~$30.0$  &$5.0$~/~$1.0$  &$91.0$~/~$26.0$   &$\underline{21.0}$~/~$2.0$  &$74.0$~/~$4.0$ &$0.0$~/~$0.0$ \\
        DeepSeek-R1-0528  & $685$B & $90.0$~/~$\underline{35.0}$ &$7.0$~/~$1.0$ &$90.0$~/~$\underline{42.0}$ & $\mathbf{28.0}$~/~$2.0$ &$76.0$~/~$\mathbf{26.0}$ &$\mathbf{14.0}$~/~$2.0$ \\
        \ours  & $8$B  &$83.0$~/~$\mathbf{36.0}$       &$\underline{10.0}$~/~$6.0$     &$\underline{97.0}$~/~$\mathbf{45.0}$   &  $17.0$~/~$0.0$   &$\underline{82.0}$~/~$\underline{20.0}$  &$\underline{10.0}$~/~$\mathbf{4.0}$ \\
        \ \ \ w/o RL (SFT only)   & $8$B &$65.0$~/~$29.0$ &$\underline{10.0}$~/~$4.0$  & $85.0$~/~$27.0$       &$8.0$~/~$\underline{3.0}$    &$64.0$~/~$6.0$  &$2.0$~/~$2.0$ \\
        \bottomrule
    \end{tabular}
    }
    \caption{Main results on \textsc{KernelBench}. 
    We present \texttt{Compilation Accuracy} (Comp), \texttt{Execution Accuracy} (Exec), $\text{fast}_\text{1}$ and $\text{fast}_\text{2}$.
    The best-performing and second-best-performing methods are highlighted in \textbf{Bold} and \underline{Underline}, respectively.
    }
    \label{tab:kernelbench_result}
\end{table*}

%% file: tables/pass_10_results.tex
\begin{table*}[ht!] \small
    \centering
    \resizebox{0.99\textwidth}{!}{
        \begin{tabular}{lcccccccccc}
        \toprule

        \multirow{2}{*}{\textbf{Model}} 
        & \multirow{2}{*}{\textbf{Lang.}} 
        & \multirow{2}{*}{\textbf{\#Params}} 
        & \multicolumn{2}{c}{\textbf{\textsc{Level1}}} & \multicolumn{2}{c}{\textbf{\textsc{Level2}}}  & \multicolumn{2}{c}{\textbf{\textsc{Level3}}} \\
        \cmidrule(lr){4-5}
        \cmidrule(lr){6-7}
        \cmidrule(lr){8-9}
        & & &  \textbf{Comp}~/~\textbf{Exec} & \textbf{P75}~/~\textbf{P50} & \textbf{Comp}~/~\textbf{Exec} & \textbf{P75}~/~\textbf{P50} &\textbf{Comp}~/~\textbf{Exec} & \textbf{P75}~/~\textbf{P50}
        \\         
        \midrule
        AI Cuda Engineer \\
        \ \ \ - \ o1-preview  & CUDA & -  &$-$~/~$63.0$ &$0.96$~/~$0.45$ & $-$~/~$\underline{95.0}$       &$1.01$~/~$1.00$    &$-$~/~$19.0$  &$1.00$~/~$0.99$ \\
        \ \ \ - \ o1-high  & CUDA & -  &$-$~/~$50.0$ &$0.97$~/~$0.37$ & $-$~/~$81.0$       &$1.00$~/~$0.87$    &$-$~/~$12.0$  &$1.00$~/~$0.93$ \\
        Claude-4-Sonnet & CUDA   & - &$99.0$~/~$64.0$ &$\mathbf{1.26}$~/~$\mathbf{0.97}$  & $\mathbf{100.0}$~/~$92.0$  &$1.42$~/~$1.19$   &$\mathbf{100.0}$~/~$66.0$  &$\mathbf{1.22}$~/~$\mathbf{1.00}$ \\
        DeepSeek-R1-0528 & CUDA  & $685$B &$99.0$~/~$\mathbf{97.0}$ &$\underline{1.23}$~/~$0.85$  & $\mathbf{100.0}$~/~$\mathbf{100.0}$       &$\mathbf{1.74}$~/~$\mathbf{1.33}$    &$\mathbf{100.0}$~/~$\underline{70.0}$  &$\underline{1.17}$~/~$\mathbf{1.00}$ \\
        Kevin* & CUDA  & $32$B &$\mathbf{100.0}$~/~$\underline{88.0}$ &$1.14$~/~$\underline{0.78}$  & $98.0$~/~$86.0$       &$\underline{1.64}$~/~$1.24$    &$\mathbf{100.0}$~/~$\underline{70.0}$  &$1.10$~/~$0.92$ \\
        KernelLLM & Triton & $8$B & $99.0$~/~$52.0$ & $-$~/~$-$ & $97.0$~/~$34.0$ & $-$~/~$-$ & $-$~/~$-$ & $-$~/~$-$ \\
        Claude-4-Sonnet & Triton  & - &$99.0$~/~$57.0$ &$1.01$~/~$0.76$  & $\mathbf{100.0}$~/~$68.0$       &$1.41$~/~$1.12$     &$99.0$~/~$60.0$  &$1.12$~/~$\mathbf{1.00}$ \\
        DeepSeek-R1-0528 & Triton  & $685$B &$\mathbf{100.0}$~/~$74.0$ &$1.04$~/~$0.62$  & $\mathbf{100.0}$~/~$74.0$       &$1.56$~/~$\underline{1.28}$ &$\mathbf{100.0}$~/~$\mathbf{72.0}$  &$1.03$~/~$0.92$ \\
        \ours  & Triton & $8$B &$\mathbf{100.0}$~/~$68.0$ &$1.01$~/~$0.69$  & $\mathbf{100.0}$~/~$88.0$       &$1.17$~/~$1.01$     &$88.0$~/~$52.0$  &$1.03$~/~$\mathbf{1.00}$ \\
        \bottomrule
    \end{tabular}
    }
    \caption{Cross comparison results for Triton and CUDA models on \textsc{KernelBench}. We report pass@10 results for each model. 
    We present \texttt{Compilation Accuracy} (Comp), \texttt{Execution Accuracy} (Exec), \texttt{Torch P75} (P75) and \texttt{Torch P50} (P50).
    The best-performing and second-best-performing methods are highlighted in \textbf{Bold} and \underline{Underline}, respectively. * denotes that they use $180$ of the evaluation data for training purpose.
    }
    \label{tab:pass@10_result}
\end{table*}

%% file: tables/reward_design_role.tex
\begin{table}[]
    \centering
    \small
    \resizebox{0.75\textwidth}{!}{
    \begin{tabular}{lcc}
        \toprule
        \textbf{Model} & \textbf{\textsc{TritonBench-T}} & \textbf{\textsc{KernelBench}-Level 1}\\
        \midrule
        \ours & $5$ & $6$ \\
        \ \ \ w/o rule-based reward & $18$ & $25$ \\
        \ \ \ w/o RL (SFT only) & $4$ & $4$ \\
        \ \ \ w/o SFT\&RL (Backbone model) & $66$ & $10$ \\

         \bottomrule
    \end{tabular}
    }
    \caption{Numbers of generated Triton code that do not contains keyword "\texttt{@triton.jit}".}
    \label{tab:reward_design_role}
\end{table}